\documentclass[3p,times]{elsarticle}

\usepackage{ecrc}
\usepackage{units}
\usepackage{import}
\usepackage{amsmath,amssymb,amsfonts}
\usepackage{tabularx}
\usepackage{caption}
\usepackage{booktabs}
\usepackage{multirow}
\usepackage[colorinlistoftodos, backgroundcolor=teal!40, linecolor=black, textwidth=18mm, textsize=scriptsize]{todonotes}
\def\expCollisionInsideROMrobotContactReachedResVal{0.90}

\def\expCollisionInsideROMrobotContactReachedRelImprov{88}

\def\expCollisionInsideROMarccvIContactReachedResVal{0.33}

\def\expCollisionInsideROMarccvIContactReachedRelImprov{68}

\def\expCollisionInsideROMarccvIIContactReachedResVal{0.11}

\def\expCollisionInsideROMrobotForceReachedResVal{3.55}

\def\expCollisionInsideROMrobotForceReachedRelImprov{63}

\def\expCollisionInsideROMarccvIForceReachedResVal{1.69}

\def\expCollisionInsideROMarccvIForceReachedRelImprov{22}

\def\expCollisionInsideROMarccvIIForceReachedResVal{1.32}

\def\expCollisionInsideROMarccvIZeroPosResVal{1.41}

\def\expCollisionInsideROMarccvIZeroPosRelImprov{64}

\def\expCollisionInsideROMarccvIIZeroPosResVal{0.51}

\def\expCollisionInsideROMrobotRMSEForceResVal{5.63}

\def\expCollisionInsideROMrobotRMSEForceRelImprov{63}

\def\expCollisionInsideROMarccvIRMSEForceResVal{2.85}

\def\expCollisionInsideROMarccvIRMSEForceRelImprov{28}

\def\expCollisionInsideROMarccvIIRMSEForceResVal{2.06}

\def\expCollisionOutsideROMrobotContactReachedResVal{5.31}

\def\expCollisionOutsideROMrobotContactReachedRelImprov{74}

\def\expCollisionOutsideROMarccvIContactReachedResVal{2.46}

\def\expCollisionOutsideROMarccvIContactReachedRelImprov{43}

\def\expCollisionOutsideROMarccvIIContactReachedResVal{1.40}

\def\expCollisionOutsideROMrobotForceReachedResVal{8.14}

\def\expCollisionOutsideROMrobotForceReachedRelImprov{66}

\def\expCollisionOutsideROMarccvIForceReachedResVal{4.31}

\def\expCollisionOutsideROMarccvIForceReachedRelImprov{35}

\def\expCollisionOutsideROMarccvIIForceReachedResVal{2.79}

\def\expCollisionOutsideROMarccvIZeroPosResVal{3.96}

\def\expCollisionOutsideROMarccvIZeroPosRelImprov{53}

\def\expCollisionOutsideROMarccvIIZeroPosResVal{1.87}

\def\expCollisionOutsideROMrobotRMSEForceResVal{12.92}

\def\expCollisionOutsideROMrobotRMSEForceRelImprov{47}

\def\expCollisionOutsideROMarccvIRMSEForceResVal{9.30}

\def\expCollisionOutsideROMarccvIRMSEForceRelImprov{26}

\def\expCollisionOutsideROMarccvIIRMSEForceResVal{6.86}

\def\expforceTrajectoryRIRrobotMaxErrorResVal{8.42}

\def\expforceTrajectoryRIRrobotMaxErrorRelImprov{82}

\def\expforceTrajectoryRIRarccvIMaxErrorResVal{1.25}

\def\expforceTrajectoryRIRarccvIMaxErrorRelImprov{-21}

\def\expforceTrajectoryRIRarccvIIMaxErrorResVal{1.51}

\def\expforceTrajectoryRIRrobotRMSEForceResVal{1.89}

\def\expforceTrajectoryRIRrobotRMSEForceRelImprov{81}

\def\expforceTrajectoryRIRarccvIRMSEForceResVal{0.22}

\def\expforceTrajectoryRIRarccvIRMSEForceRelImprov{-61}

\def\expforceTrajectoryRIRarccvIIRMSEForceResVal{0.36}

\def\expforceTrajectoryRIRarccvIRMSEPosResVal{5.19}
\def\expforceTrajectoryRIRarccvIRMSEPosResExp{-5}
\def\expforceTrajectoryRIRarccvIRMSEPosRelImprov{34}

\def\expforceTrajectoryRIRarccvIIRMSEPosResVal{3.43}
\def\expforceTrajectoryRIRarccvIIRMSEPosResExp{-5}

\def\expforceTrajectoryChirprobotCutoffResVal{1.39}

\def\expforceTrajectoryChirprobotCutoffRelImprov{1150}

\def\expforceTrajectoryChirparccvICutoffResVal{8.13}

\def\expforceTrajectoryChirparccvICutoffRelImprov{113}

\def\expforceTrajectoryChirparccvIICutoffResVal{17.34}

\def\exppihrobottimeResVal{29.77}

\def\exppihrobottimeRelImprov{78}

\def\exppiharccvItimeResVal{19.35}

\def\exppiharccvItimeRelImprov{66}

\def\exppiharccvIItimeResVal{6.56}

\def\expgearrobottimeAssemblyResVal{138.50}

\def\expgearrobottimeAssemblyRelImprov{54}

\def\expgeararccvItimeAssemblyResVal{85.10}

\def\expgeararccvItimeAssemblyRelImprov{25}

\def\expgeararccvIItimeAssemblyResVal{63.50}

\def\expgearrobottimeGearIResVal{37.00}

\def\expgearrobottimeGearIRelImprov{72}

\def\expgeararccvItimeGearIResVal{19.30}

\def\expgeararccvItimeGearIRelImprov{46}

\def\expgeararccvIItimeGearIResVal{10.40}

\def\expgearrobottimeGearIIResVal{62.00}

\def\expgearrobottimeGearIIRelImprov{77}

\def\expgeararccvItimeGearIIResVal{27.70}

\def\expgeararccvItimeGearIIRelImprov{49}

\def\expgeararccvIItimeGearIIResVal{14.10}

\def\expboardrobottimeAssemblyResVal{40.00}

\def\expboardrobottimeAssemblyRelImprov{33}

\def\expboardarccvItimeAssemblyResVal{27.90}

\def\expboardarccvItimeAssemblyRelImprov{4}

\def\expboardarccvIItimeAssemblyResVal{26.70}

\def\expboardrobottimeBoardResVal{19.10}

\def\expboardrobottimeBoardRelImprov{70}

\def\expboardarccvItimeBoardResVal{6.90}

\def\expboardarccvItimeBoardRelImprov{17}

\def\expboardarccvIItimeBoardResVal{5.70}

\volume{00}

\firstpage{1}

\journalname{Robotics and Computer-Integrated Manufacturing}

\runauth{}

\jid{RCOM}

\jnltitlelogo{RCIM}

\CopyrightLine{2026}{Published by Elsevier Ltd.}

\usepackage{amssymb}
\usepackage[figuresright]{rotating}

\begin{document}

\begin{frontmatter}

\dochead{}

\title{A Unified Control Architecture for Macro-Micro Manipulation using a Active Remote Center of Compliance for Manufacturing Applications}

\author[label2]{Patrick Frank\corref{cor1}}
\ead{patrick.frank@h-ka.de}
\cortext[cor1]{Corresponding author}
\author[label2]{Christian Friedrich}

\address[label2]{Institute for Robotics and Intelligent Production Systems\\ University of Applied Sciences Karlsruhe (HKA)\\Moltkestraße 30, 76133 Karlsruhe, Germany.}

\begin{abstract}
Macro-micro manipulators combine a macro manipulator with a large workspace, such as an industrial robot, with a lightweight, high-bandwidth micro manipulator. This enables highly dynamic interaction control while preserving the wide workspace of the robot. Traditionally, position control is assigned to the macro manipulator, while the micro manipulator handles the interaction with the environment, limiting the achievable interaction control bandwidth. To solve this, we propose a novel control architecture that incorporates the macro manipulator into the active interaction control. This leads to a increase in control bandwidth by a factor of 2.1 compared to the state of the art architecture, based on the leader-follower approach and factor 12.5 compared to traditional robot-based force control. Further we propose surrogate models for a more efficient controller design and easy adaptation to hardware changes. We validate our approach by comparing it against the other control schemes in different experiments, like collision with an object, following a force trajectory and industrial assembly tasks.
\end{abstract}

 
\end{frontmatter}

\section{Introduction}
\label{sec:introduction}
\begin{figure}
\centering
\includegraphics[width=\linewidth]{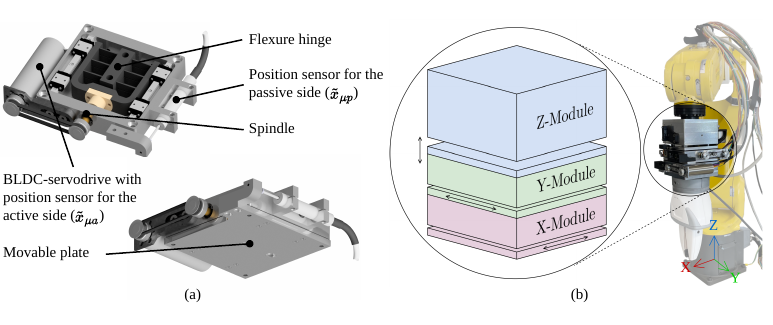}
\caption{(a) Enhanced module of the Active Remote Center of Compliance (ARCC) from \cite{friedrichHighlyDynamicPhysical2025}. Each module consists of a BLDC-servodrive, a spindle, a flexure hinge and a position sensor to measure the spring deflection. (b) Configuration with three linear modules mounted on a industrial robot.}
\label{fig:arccOverview}
\end{figure}

Dynamic physical interaction is crucial in robotic tasks like assembly/disassembly, grinding and polishing, where both sensitivity and low cycle times are required. Control strategies for such scenarios are generally classified as passive or active \cite{villaniForceControl2016}. Passive methods, such as using compliant devices like the Remote Center of Compliance (RCC) \cite{whitneyMechanicalBehaviorDesign1986a, watsonRemoteCenterCompliance1978a}, are cost-effective but inflexible, as they are tailored to specific tasks. Active interaction control, including impedance/admittance control \cite{hoganImpedanceControlApproach1984, petitCartesianImpedanceControl2011a, iskandarHybridForceImpedanceControl2023c}, and hybrid \cite{raibertHybridPositionForce1981} or parallel \cite{langeForceTrajectoryControl2013a} force/position approaches, overcomes this limitation by adapting control parameters to different tasks \cite{schemppPIPEProcessInformed2024a, weiContactForceEstimation2024a}. However, these methods typically suffer from lower dynamic performance due to the robots high inertia. 
This can be solved by using active end-effectors employed in a macro-micro approach \cite{sharonEnhancementRobotAccuracy1984}. In this configuration, an industrial robot (macro manipulator) positions the end-effector and provides a large workspace and flexibility. A smaller, high-performance active end-effector (micro manipulator) is responsible for fine positioning, precise force control, and high-bandwidth responses. To further enhance the performance of the contact interaction, passive compliance elements, such as flexure hinges, based on the principle of a series elastic actuator \cite{prattSeriesElasticActuators1995} can be incorporated between the servo drive and the load. These systems are sometimes also reffered to as Active Remote Center of Compliance (ARCC) \cite{friedrichHighlyDynamicPhysical2025}, see Fig. \ref{fig:arccOverview} (a).

\section{Related Work}
\label{sec:relatedWork}
Most of the macro-micro manipulation approaches are used in the field of grinding and polishing applications.
Wu et al. \cite{wuAdaptiveNeuralNetwork2015} employ a adaptive neural network compensator to decouple the micro from the macro dynamics.
Fan Chen et al. \cite{chenContactForceControl2019} control the contact force in a polishing application by an active end-effector with one degree of freedom (DoF) as the micro and a industrial robot as the macro manipulator.
In their later work \cite{chenRoboticGrindingBlisk2019a} they added a second DoF to the end-effector and extended the control architecture by a position controller to keep the tool in its center position tangential to the workpiece during grinding.
Ma et al. \cite{maDesignControlEndeffector2018} propose to employ valve position control \cite{luybenEssentialsProcessControl1997} to ensure that their three DoF micro manipulator returns to its center position in steady state. In this position, the micro manipulator has the largest range of motion for responding to disturbances. However, they didn't implement this into their control framework. Therefore the macro and micro manipulator work independently.
Arifin et al. \cite{arifinGeneralFrameworkForce2013a} propose a framework for macro-micro control where the macro manipulator ensures that the micro manipulator stays in its center position.
Friedrich et al.\cite{friedrichHighlyDynamicPhysical2025} employ a similar control approach for their one DoF micro manipulator and use it for contour following, peg-in-hole and assembly tasks.
Haiqing Chen et al. \cite{chenRoboticCompliantGrinding2024a} designed a active one DoF end-effector for grinding applications where the macro manipulator tracks a position trajectory while the micro performs force control.
In their later work \cite{chenRoboticGrindingCurved2025} they used a two DoF end-effector and enhanced the control architecture by adding a force-position decoupling algorithm to improve the control accuracy of the normal force and tangential tool displacement. Chin-Yin Chen et al. \cite{chenSensorbasedForceDecouple2023a} propose a dynamic decoupling method for macro-micro manipulations by employing two force sensors in their active end-effector.
Zhou et al. \cite{zhouFinitetimeSMCbasedAdmittance2025} employ sliding mode control as a adaptive input into the admittance controller to improve tracking of the polishing force and therefore the surface quality.
Lopes and Almeida \cite{lopesForceImpedanceControlled2008} use a six DoF parallel micro-manipulator with an impedance task-space controller and an outer force controller for contour following and peg-in-hole tasks.

All approaches have in common, that they follow one of two well-known control principles. The first one considers the manipulators seperately, meaning there is no feedback between the macro and micro manipulator. The macro manipulator is position controlled, while the micro manipulator is force controlled, see \cite{chenContactForceControl2019, chenRoboticGrindingBlisk2019a, chenRoboticCompliantGrinding2024a, chenRoboticGrindingCurved2025, zhouFinitetimeSMCbasedAdmittance2025, heDesignNovelForce2023, daiMacroMiniManipulatorBased2024b}. As long as the required movement of the micro manipulator is within its range of motion (ROM) this is a valid approach. For use-cases where the micro manipulators ROM isn't enough, the second control principle is used. This is often reffered as a leader-follower-based approach (LF). Here, the macro manipulators objective is to keep the micro manipulator in its center position
\cite{friedrichHighlyDynamicPhysical2025, arifinGeneralFrameworkForce2013a, chenSensorbasedForceDecouple2023a}. The drawback of this principle is, that the macro manipulator isn't a part of the active force control, limiting the achievable force control bandwidth. 

Our proposed control architecture is motivated by this drawback. We incorporate both, the macro and micro manipulator, in the active force control resulting in a significant dynamics increase.
The main contributions of this paper are:
\begin{itemize}
\item Novel control architecture for macro-micro manipulation where both, the macro and micro, are part of the active interaction control, leading to a significant increase in control bandwidth by a factor of 2.1 compared to the state of the art LF-based approach.
\item Derivation and identification of surrogate models for the macro-micro manipulator to allow for a more efficient controller design and easy adaption to hardware changes in a modular way.
\item Validation on different experiments and real-world tasks to compare our architecture against approaches like LF-based and robot-based force control (RB).
\end{itemize}

\section{Control Design}
\label{sec:controlDesign}
For the remainder of this paper we use a industrial robot (FANUC LR Mate 200iD/7L) as the macro manipulator and a updated version of the active remote center of compliance (ARCC) from \cite{friedrichHighlyDynamicPhysical2025} as the micro. The ARCC consists of a flexure hinge for passive compliance, a BLDC-servodrive for active compliance and a position measuring system. We use three linear modules in a Cartesian configuration as shown in Fig. \ref{fig:arccOverview} to cover the three linear DoF. First, we provide different surrogate models for the ARCC and robot in section \ref{subsec:ControlDesignModeling}. Based on this, we demonstrate in section \ref{subsec:ControlDesignSystemIdentification} their effectiveness and accuracy, using classic system identification strategies. This forms the foundation of our control architecture (section \ref{subsec:ControlDesignArchitecture}) and controller synthesis (section \ref{subsec:ControlDesignSynthesis}).
\subsection{Modeling}
\label{subsec:ControlDesignModeling}
\begin{figure}[!h]
\fontsize{6pt}{7pt}\selectfont
\centering
\def\svgwidth{\linewidth}
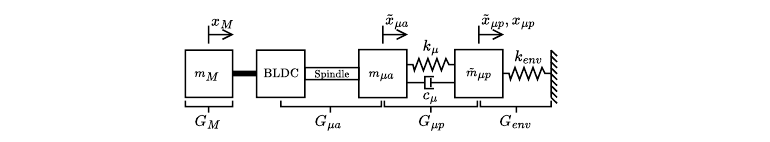
\caption{Mechanical model of the macro-micro manipulator with the corresponding transfer functions. Index $M$ denotes the macro manipulator, index $\mu$ the micro manipulator with the additional indices $a$ for the active part and $p$ for the passive part, respectively.}
\label{fig:mechanicalModel}
\end{figure}

The macro-micro manipulator and its interaction with the environment can be described using a mass-spring-damper system, see Fig. \ref{fig:mechanicalModel}. For simplicity we only use one DoF here. The macro manipulator, a industrial robot, is labeled by index $M$, the micro manipulator by index $\mu$. 

\noindent For the macro manipulator the transfer function $G_M$ (\ref{eq:tfMacro}) in Cartesian space from the desired velocity $\dot{x}_{M,des}$ to the actual velocity $\dot{x}_{M,act}$ in a small subworkspace can be approximated by a second-order low-pass. Here, $K_M$ is a gain factor, $\omega_{cM}$ is a cutoff frequency and $\zeta_{M}$ a damping ration to be identified. 
\begin{equation}
G_M(s) = \frac{\dot{X}_{M,act}(s)}{\dot{X}_{M,des}(s)} = \frac{K_M\cdot \omega_{cM}^2}{s^2 + 2\cdot\zeta_{M}\cdot \omega_{cM} \cdot s + \omega_{cM}^2} \label{eq:tfMacro}
\end{equation}

\noindent The micros drive side, consisting of a BLDC-servodrive and a spindle, is called the \textit{active side} and is labeled by index $a$. The active side is connected to the robot by the ARCCs housing. The index $p$ labels the \textit{passive side}, i.e. end-effector flange of the ARCC. The stiffness and damping of the flexure hinge are denoted by $k_{\mu}$ and $c_{\mu}$, respectively. The interaction with the environment can be described by a spring with stiffness $k_{env}$. From Fig. \ref{fig:mechanicalModel} we can derive the equation of motion (\ref{eq:equationsOfMotion}).
\begin{equation}
\tilde{m}_{\mu p}\cdot \ddot{x}_{\mu p}(t)  = - k_{env}\cdot x_{\mu p}(t) -c_\mu\cdot (\dot{\tilde{x}}_{\mu p}(t) - \dot{\tilde{x}}_{\mu a}(t))  - k_\mu\cdot (\tilde{x}_{\mu p}(t) - \tilde{x}_{\mu a}(t))  \label{eq:equationsOfMotion}
\end{equation}
where $\tilde{x}_{\mu a} = x_{\mu a} - x_M$ and $\tilde{x}_{\mu p} = x_{\mu p} - x_M$ are the positions of the active and passive side inside the tool center point frame, respectively. The total end-effector mass, i.e. passive side, gripper and payload, is described by $\tilde{m}_{\mu p} = m_{\mu p} + m_{load}$.

The internal motion control of the ARCCs BLDC-servodrive is realized by a P-PI-PI cascade. The PI torque control is tuned using the magnitude optimum and the symmetrical optimum is used for the PI velocity control, see \cite{ElektrischeAntriebeRegelung2009}. Following this approach the transmission behavior of the desired velocity $\dot{\tilde{x}}_{\mu a,des}$ to the actual velocity $\dot{\tilde{x}}_{\mu a,act}$ of the active side can be described by a second-order low-pass (\ref{eq:tfMicroActive}).
\begin{equation}
G_{\mu a}(s) = \frac{\dot{\tilde{X}}_{\mu a,act}(s)}{\dot{\tilde{X}}_{\mu a,des}(s)} = \frac{K_{\mu a}\cdot \omega_{c\mu a}^2}{s^2 + 2\cdot \zeta_{\mu a}\cdot \omega_{c\mu a} \cdot s + \omega_{c\mu a}^2} \label{eq:tfMicroActive}
\end{equation}
where $K_{\mu a}$ is a gain factor, $\zeta_{\mu a}$ a damping ratio and $\omega_{c\mu a}$ a cutoff frequency to be identified.
The transfer function $G_{\mu p}$ (\ref{eq:tfMicroPassive}) from the micro manipulators active to passive side is obtained by transforming (\ref{eq:equationsOfMotion}) into the frequency domain with $s = \sigma + j\cdot \omega$. $x_M$ is set to $0$ to exclude the influence of the macro manipulator.
\begin{equation}
G_{\mu p}(s) = \frac{\tilde{X}_{\mu p}(s)}{\tilde{X}_{\mu a}(s)} = \frac{c_\mu\cdot s + k_\mu}{\tilde{m}_{\mu p}\cdot s^2 + c_\mu\cdot s + (k_\mu + k_{env})} \label{eq:tfMicroPassive}
\end{equation}
 From (\ref{eq:tfMicroActive}) and (\ref{eq:tfMicroPassive}) we derive the complete transfer function (\ref{eq:tfMicro}) of the micro manipulator $G_\mu$.

\begin{equation}
G_{\mu}(s) = G_{\mu p}(s)\cdot \frac{1}{s}\cdot G_{\mu a}(s) = \frac{\tilde{X}_{\mu p}(s)}{\dot{\tilde{X}}_{\mu a,des}(s)} =\frac{K_{\mu a}\cdot \omega_{c\mu a}^2\cdot (c_\mu{\cdot} s + k_\mu)}{(s^2 + 2\cdot \zeta_{\mu a}\cdot \omega_{c\mu a} \cdot s + \omega_{c\mu a}^2)\cdot s\cdot(\tilde{m}_{\mu p}\cdot s^2 + c_\mu\cdot s + (k_\mu + k_{env}))} \label{eq:tfMicro}
\end{equation}

\noindent Lastly, the transfer function for the interaction with the environment $G_{env}$ is given by (\ref{eq:tfEnvironment}).
\begin{equation}
G_{env}(s) = \frac{F_{act}(s)}{X_{\mu p}(s)} = k_{env} \label{eq:tfEnvironment}
\end{equation}

\subsection{System Identification}
\label{subsec:ControlDesignSystemIdentification}
The parameters for the transfer functions (\ref{eq:tfMacro}), (\ref{eq:tfMicroActive}) and (\ref{eq:tfMicroPassive}) are identified using a parametric approach based on the instrument variable algorithm. A sine sweep is applied as the identification signal with $f_{id,M}=[1\dots 20]\,\mathrm{Hz}$ and $f_{id,\mu}=[5\dots 100]\,\mathrm{Hz}$ for the macro and micro manipulator, respectively, with an amplitude of $A_{\dot{x}} = \unitfrac[0.1]{m}{s}$. 

\textit{Macro manipulator}: The macro manipulator is identified at equally distributed positions inside the expected workspace as shown in Fig. \ref{fig:boxplotMacro} (a). For the three linear DoFs there is very little variation in the transmission behavior inside the workspace, as shown in Fig. \ref{fig:boxplotMacro} (b). The identified parameters are listed in Tab. \ref{tab:resSysidentMacro}. \\
\begin{table}[h]
\caption{Results of the system identification of the macro manipulators three linear DoFs.}
\label{tab:resSysidentMacro}
\centering
\begin{tabular}{lccccc}
\toprule
 & $K_{M}$ & $\zeta_M$ & $\omega_{cM}$ [Hz] & Model Fit [\%] & MSE [\nicefrac{m}{s}] \\
\midrule
X-axis & 1.25 $\pm$ 4.61$\cdot 10^{-3}$ & 1 & 3.20 $\pm$ 1.33$\cdot 10^{-2}$ & 91.6 $\pm$ 0.6 & 1.37$\cdot 10^{-5}$ $\pm$ 2.06$\cdot 10^{-6}$\\
Y-axis & 1.26 $\pm$ 2.57$\cdot 10^{-3}$ & 1 & 3.19 $\pm$ 1.26$\cdot 10^{-2}$ & 91.6 $\pm$ 0.4 & 1.35$\cdot 10^{-5}$ $\pm$ 1.45$\cdot 10^{-6}$\\
Z-axis & 1.25 $\pm$ 2.22$\cdot 10^{-3}$ & 1 & 3.19 $\pm$ 1.10$\cdot 10^{-2}$ & 91.6 $\pm$ 0.4 & 1.35$\cdot 10^{-5}$ $\pm$ 1.31$\cdot 10^{-6}$ \\
\bottomrule
\end{tabular}
\end{table} \pagebreak

\noindent\begin{minipage}{\linewidth}
\begin{minipage}[c]{\linewidth}
\includegraphics[width=\linewidth]{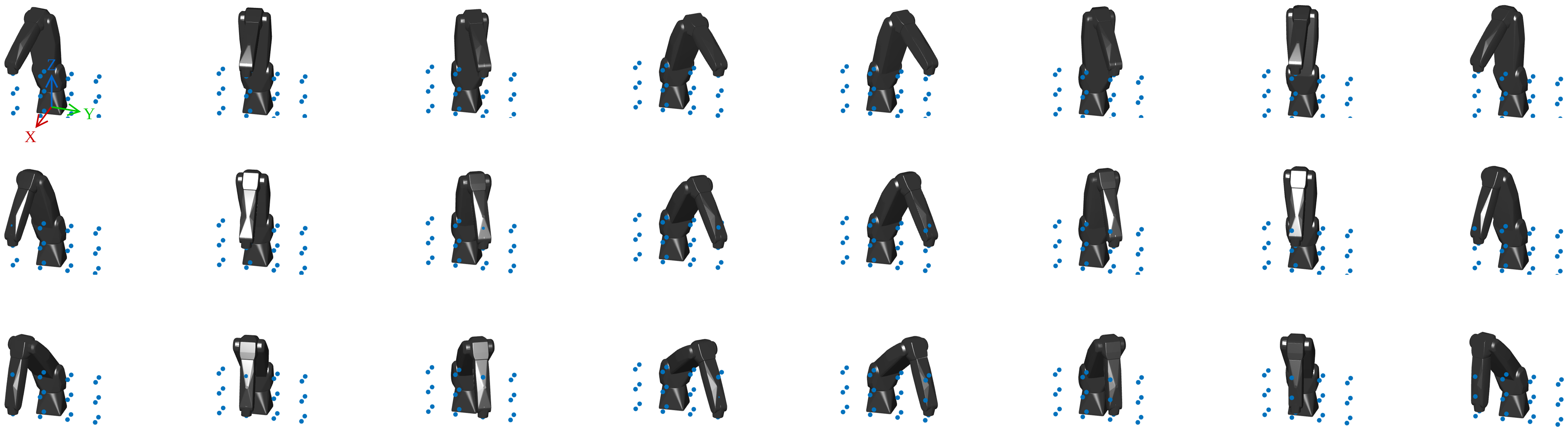}
\begin{center}
    \vspace*{-0.25cm}(a)\vspace*{0.25cm}
\end{center}
\end{minipage}\\
\begin{minipage}[c]{\linewidth}
\includegraphics[width=\linewidth]{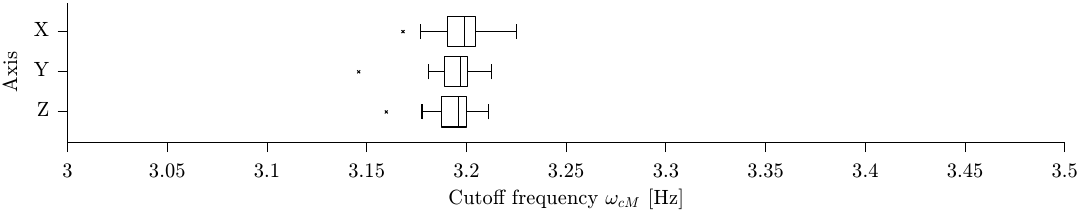}
\begin{center}
    \vspace*{-0.25cm}(b)
\end{center}
\end{minipage}\\
\captionof{figure}{(a) Sampling positions for the system identification inside the expected workspace ($\mathrm{X{\times} Y{\times} Z}$) of $\unit[200{\times} 600{\times} 300]{mm}$. (b) Resulting boxplot of the identified cutoff frequencies $\omega_{cM}$ for the three linear DoFs of the macro manipulator.}
\label{fig:boxplotMacro}
\end{minipage}\\

\begin{figure*}[hb!]
\centering
\includegraphics[width=\textwidth, keepaspectratio]{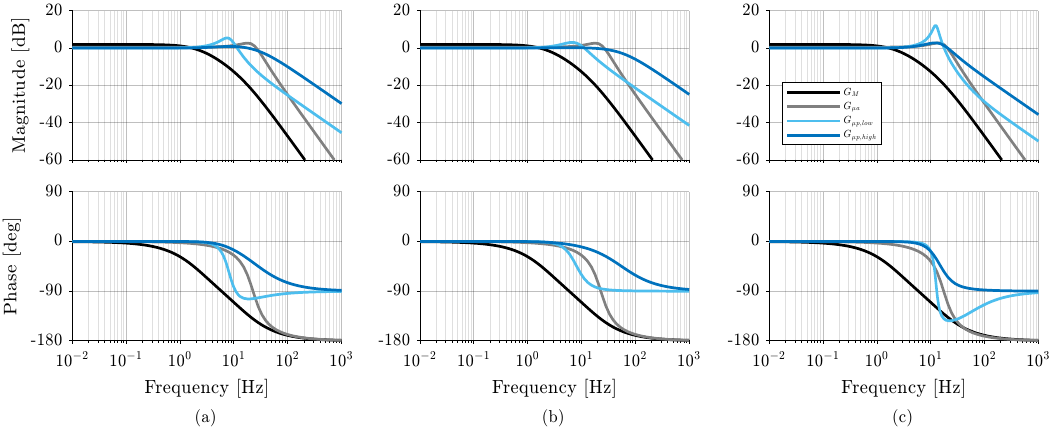}
\caption{Bode plots of the identified transfer functions for the macro manipulator $G_M$, the micro manipulators active side $G_{\mu a}$ and passive side $G_{\mu p}$. The additional indices $low$ and $high$ denote the low stiffness and high stiffness flexure hinge, respectively. (a) X-axis with $k_{\mu, low}=15\;\unitfrac{N}{mm}$ and $k_{\mu, high}=30\;\unitfrac{N}{mm}$, (b) Y-axis with $k_{\mu, low}=15\;\unitfrac{N}{mm}$ and $k_{\mu, high}=30\;\unitfrac{N}{mm}$, (c) Z-axis with $k_{\mu, low}=20\;\unitfrac{N}{mm}$ and $k_{\mu, high}=40\;\unitfrac{N}{mm}$.}
\label{fig:resultSysidentBode}
\end{figure*}

\textit{Micro manipulator}: The results for the system identification of the micro manipulators active side are listed in Tab. \ref{tab:resSysidentMicroActiveSide}. The passive side of the micro manipulator is identified with a payload of $\unit[1.5]{kg}$ and for two different flexure hinge stiffnesses. The low stiffness flexure hinge has a stiffness of $k_{\mu ,low} = \unitfrac[15]{N}{mm}$ for the X- and Y-axis and $k_{\mu ,low} = \unitfrac[20]{N}{mm}$ for the Z-axis. The high stiffness flexure hinge has a stiffness of $k_{\mu ,high} = \unitfrac[30]{N}{mm}$ for the X- and Y-axis and $k_{\mu ,high} = \unitfrac[40]{N}{mm}$ for the Z-axis. The frequency responses of the identified transfer functions (\ref{eq:tfMacro}), (\ref{eq:tfMicroActive}) and (\ref{eq:tfMicroPassive}) are shown in Fig. \ref{fig:resultSysidentBode}.

\pagebreak
\begin{table}[h]
\caption{Results of the system identification of the micro manipulators active side for all three linear DoFs.}
\label{tab:resSysidentMicroActiveSide}
\centering
\begin{tabular}{lccccc}
\toprule
 & $K_{\mu a}$ & $\zeta_{\mu a}$ & $\omega_{c\mu}$ [Hz] & Model Fit [\%] & MSE [\nicefrac{m}{s}] \\
\midrule
X-axis & 1.29 & 0.45 & 26.87 & 94.4 & 1.36$\cdot 10^{-5}$ \\
Y-axis & 1.30 & 0.45 & 27.05 & 94.1 & 1.53$\cdot 10^{-5}$ \\
Z-axis & 1.30 & 0.42 & 20.71 & 96.0 & 6.73$\cdot 10^{-6}$ \\
\bottomrule
\end{tabular}
\end{table}

\textit{Environment}: The stiffness of the environment $k_{env}$ is identified at different positions in the workspace by pushing the end-effector against a rigid object and measuring the force using a force-torque sensor (FTS) and the position of the end-effector. This stiffness also includes the compliance of the robot arm. The measured stiffness values are shown in Fig. \ref{fig:measurementEnvironment}. The force controlled part of our experiments is located around $X = \unit[0.45]{m}$ and $Y = \unit[0.1]{m}$, therefore we will use the stiffnesse values $k_{env, x} = \unit[40]{\nicefrac{N}{mm}}$, $k_{env, y} = \unit[30]{\nicefrac{N}{mm}}$ and $k_{env, z} = \unit[60]{\nicefrac{N}{mm}}$ for the X-, Y- and Z-axis, respectively  \\

\noindent\begin{minipage}{\linewidth}
\begin{minipage}[c]{\linewidth}
\includegraphics[width=\linewidth]{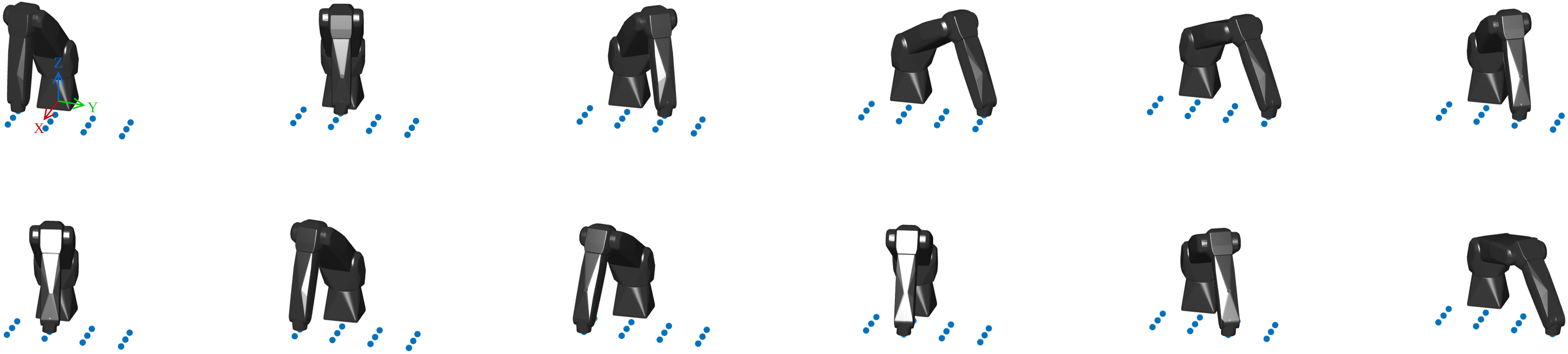}
\begin{center}
(a)\vspace*{0.5cm}
\end{center}
\end{minipage}\\
\begin{minipage}[c]{\linewidth}
\includegraphics[width=\linewidth]{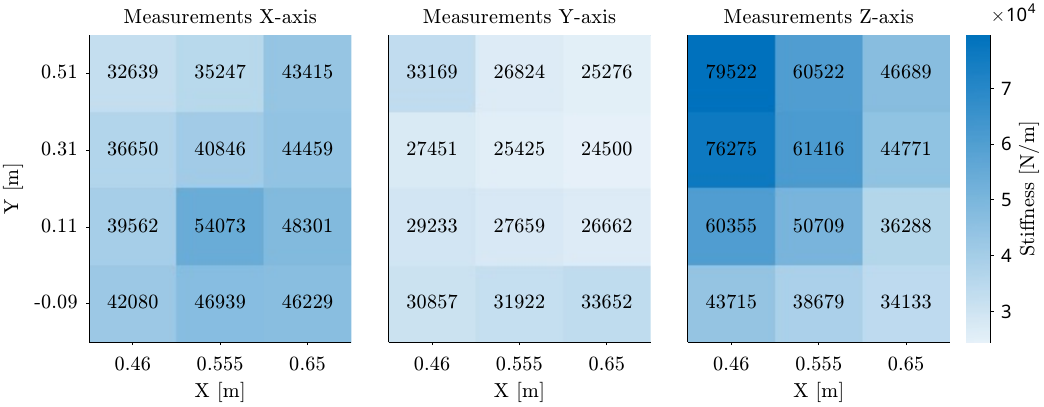}
\hspace*{3.25cm}(b) \hspace*{4.2cm} (c) \hspace*{4.2cm} (d)
\end{minipage}\\
\captionof{figure}{(a) Sampling positions inside the expected workspace for the measurement of the environment stiffness $k_{env}$ used in the surrogate model described by (\ref{eq:tfEnvironment}). (b), (c), (d) Measured stiffnesses for the X-axis, Y-axis and Z-axis, respectively.}
\label{fig:measurementEnvironment}
\end{minipage}\\

\subsection{Control Architecture}
\label{subsec:ControlDesignArchitecture}

\begin{figure}[h!]
\fontsize{6pt}{7pt}\selectfont
\centering
\def\svgwidth{\linewidth}
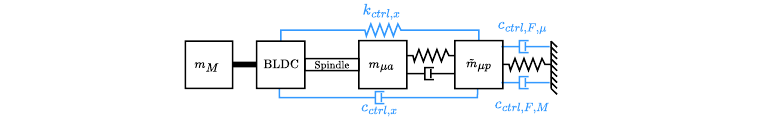
\caption{Virtual spring and damper elements to shape the dynamics of the closed loop system.}
\label{fig:mechanicalModelController}
\end{figure}

Based on the mechanical model in Fig. \ref{fig:mechanicalModel} we add virtual springs and dampers, as illustrated by Fig. \ref{fig:mechanicalModelController}, to shape the dynamics of the closed loop system, similar to the idea behind admittance and impedance control \cite{hoganImpedanceControlApproach1984}. The virtual dampers $c_{ctrl,F,M}$ and $c_{ctrl,F,\mu}$ are added to stabilize the interaction with the environment, the virtual spring $k_{ctrl,x,\mu}$ is used to keep the micro manipulator in its center position. Lastly, a virtual damper $c_{ctrl,x,\mu}$ is added to the macro for additional damping. This leads to two damping controllers for the force control based on the principle of admittance control. 

\begin{figure}[h!]
\fontsize{6pt}{7pt}\selectfont
\centering
\def\svgwidth{\linewidth}
\begingroup%
  \makeatletter%
  \providecommand\color[2][]{%
    \errmessage{(Inkscape) Color is used for the text in Inkscape, but the package 'color.sty' is not loaded}%
    \renewcommand\color[2][]{}%
  }%
  \providecommand\transparent[1]{%
    \errmessage{(Inkscape) Transparency is used (non-zero) for the text in Inkscape, but the package 'transparent.sty' is not loaded}%
    \renewcommand\transparent[1]{}%
  }%
  \providecommand\rotatebox[2]{#2}%
  \newcommand*\fsize{\dimexpr\f@size pt\relax}%
  \newcommand*\lineheight[1]{\fontsize{\fsize}{#1\fsize}\selectfont}%
  \ifx\svgwidth\undefined%
    \setlength{\unitlength}{396.9599762bp}%
    \ifx\svgscale\undefined%
      \relax%
    \else%
      \setlength{\unitlength}{\unitlength * \real{\svgscale}}%
    \fi%
  \else%
    \setlength{\unitlength}{\svgwidth}%
  \fi%
  \global\let\svgwidth\undefined%
  \global\let\svgscale\undefined%
  \makeatother%
  \begin{picture}(1,0.24909312)%
    \lineheight{1}%
    \setlength\tabcolsep{0pt}%
    \put(0,0){\includegraphics[width=\unitlength,page=1]{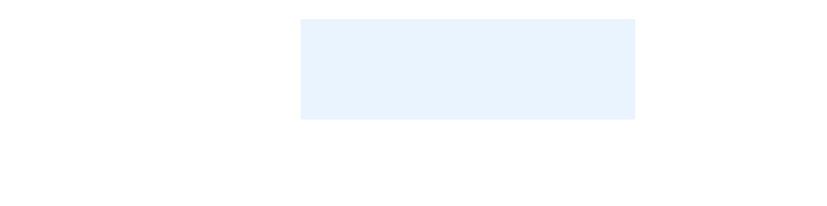}}%
    \put(0.66801169,0.21427724){\color[rgb]{0,0,0}\makebox(0,0)[lt]{\lineheight{1.25}\smash{\begin{tabular}[t]{l}Micro manipulator\end{tabular}}}}%
    \put(0,0){\includegraphics[width=\unitlength,page=2]{figure_7_architecture_arcc_v2.pdf}}%
    \put(0.66591469,0.02928768){\color[rgb]{0,0,0}\makebox(0,0)[lt]{\lineheight{1.25}\smash{\begin{tabular}[t]{l}Macro manipulator\end{tabular}}}}%
    \put(0,0){\includegraphics[width=\unitlength,page=3]{figure_7_architecture_arcc_v2.pdf}}%
  \end{picture}%
\endgroup%

\caption{Overall proposed control architecture for one DoF. The interaction planner is used to generate the position and force set points. $(\cdot)$ refers to the different equations.}
\label{fig:architectureARCCV2}
\end{figure}

\noindent For the macro manipulator the control law is given by (\ref{eq:ctrlForceMacro}):
\begin{equation}
G_{F,M}(s) = \frac{\dot{X}_{M,des}(s)}{F_{err}(s)} = \frac{1}{c_{ctrl,F,M}} \label{eq:ctrlForceMacro}
\end{equation}
\noindent where $F_{err}(s) = F_{des}(s) - F_{act}(s)$ is the force error between the desired and actual force.
Similarly, the control law for the micro manipulator is given by (\ref{eq:ctrlForceMicro}):
\begin{equation}
G_{F,\mu}(s) = \frac{\dot{\tilde{X}}_{\mu a,des,F}(s)}{F_{err}(s)} = \frac{1}{c_{ctrl,F,\mu}} \label{eq:ctrlForceMicro}
\end{equation}
The third controller is a PDT1-controller for the position error to the center position of the micro manipulator (\ref{eq:ctrlPositionMicro}):
\begin{equation}
G_{x,\mu}(s) = \frac{\dot{\tilde{X}}_{\mu a,des,x}(s)}{\tilde{X}_{\mu p,err}(s)} = k_{ctrl,x} + \frac{c_{ctrl,x}\cdot s }{T_\mathrm{filter}\cdot s + 1}\label{eq:ctrlPositionMicro}
\end{equation}
where $\tilde{X}_{\mu p,err}(s) = \tilde{X}_{\mu p,des}(s) - \tilde{X}_{\mu p,act}(s)$ is the position error between the desired and actual position of the micros passive side and $T_{\mathrm{filter}}$ is a filter time constant for the derivative term. The overall proposed control architecture is illustrated by Fig. \ref{fig:architectureARCCV2}.
Since we currently only focus on linear motions, meaning no change in orientation of the end-effector, we don't compensate for gravity. A force bias is set at the beginning of a task to compensate for the mass of the end-effector and payload.

\subsection{Controller Synthesis}
\label{subsec:ControlDesignSynthesis}
\begin{figure}[h!]
\fontsize{6pt}{7pt}\selectfont
\centering
\def\svgwidth{\linewidth}
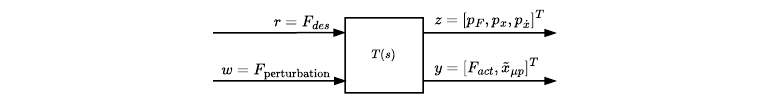
\caption{Generalized state space model $T(s)$ for the $\mathcal{H}_\infty$ synthesis. $r$: reference signal; $y$: model output; $w$: exogenous inputs, e.g. disturbances; $z$: performance functions.}
\label{fig:hinfModel}
\end{figure}

For the design of the individual controllers we use $\mathcal{H}_\infty$ synthesis for fixed-structured control systems based on the research of Apkarian et al. \cite{apkarianNonsmoothSynthesis2006} and Bruinsma et al. \cite{bruinsmaFastAlgorithmCompute1990}. We define the optimization problem as a generalized state-space model $T(s)$ with fixed and tunable components and augment it by performance functions, see Fig. \ref{fig:hinfModel}.

The signal $r$ is the reference for the model, in our case $F_{des}$. We chose not to include the reference of $\tilde{x}_{\mu p,des}$ in the optimization as it will be at 0 (center position) most of the time. The signal $y$ denotes the system output, which is $F_{act}$. Signal $w$ is for exogenous inputs, such as force disturbances. Signal $z$ is a vector of performance functions $z = [p_F, p_x, p_{\dot{x}}]^T$, which are to be minimized to meet the encoded control objectives. The first control objective is to achieve accurate force reference tracking at low frequencies and limited overshoot for high frequency changes. The second objective ensures that the micro manipulator will reach its desired position, usually the center position, once the desired force is reached. The third and last objective limits the velocity of the micro manipulators BLDC-servodrive to a feasible value.

For the performance functions $p(s)$ we define weight functions $W(s)$ for a normalized error signal $e(s)$. A performance function must statisfy $||W(s)\cdot e(s)||_\infty < 1$, compare \cite{lundstromPerformanceWeightSelection1991}. For the force tracking the weight function $W_F$ is given by (\ref{eq:HinfWF}):
\begin{equation}
    W_F = \frac{K_{hf,F}\cdot s + \omega_{co,F}}{s + \omega_{co,F}/K_{lf,F}}
    \label{eq:HinfWF}
\end{equation}
where $\omega_{co,F}$ is the crossover frequency from the low frequency gain $K_{lf,F}$ to the high frequency gain $K_{hf,F}$. $\omega_{co,F}$ is increased until $||T(s)||_\infty \geq 1$. This indicates that the system is not able to statisfy all performance goals \cite{lundstromPerformanceWeightSelection1991}. For frequencies $\omega < \omega_{co,F}$ we want good reference tracking with a error of $F_{err,des,lf} = 0.01 = \unit[1]{\%}$ of the desired force $F_{des} = \unit[20]{N}$ and therefore $K_{lf,F} = 1/F_{err,des,lf} = 100$. For frequencies $\omega > \omega_{co,F}$ a tracking error of $F_{err,des,hf} = 0.1 = \unit[10]{\%}$ is acceptable, resulting in $K_{hf,F} = 1/F_{err,des,hf} = 10$. Using this weight the performance function is defined as $p_F(s) = W_F(s)\cdot F_{err, norm}(s)$ with $F_{err, norm}(s) = F_{err}(s)/F_{des}$. Similarly, for the position error of the micro manipulator the weight $W_x$ is defined as (\ref{eq:HinfWx}).
\begin{equation}
    W_x = \frac{K_{hf,x}\cdot s + \omega_{co,x}}{s + \omega_{co,x}/K_{lf,x}}
    \label{eq:HinfWx}
\end{equation}
Here, $\omega_{co,x}$ is the crossover frequency from the low frequency gain $K_{lf,x}$ to the high frequency gain $K_{hf,x}$. For frequencies $\omega < \omega_{co,x}$ we want a low tracking error of $x_{err,des,lf} = 0.001 = \unit[0.1]{\%}$ of the micro manipulators ROM. For $\omega > \omega_{co,x}$ we relax the tracking error to $x_{err,des,hf} = 1 = \unit[100]{\%}$ to allow for large movements of the micro manipulator due to external forces, e.g. collision with an object. This results in $K_{lf,x} = 1/x_{err,des,lf} = 1000$ and $K_{hf,x} = 1/x_{err,des,hf} = 1$ for the low and high frequency gain, respectively. The performance function is defined as $p_x(s) = W_x(s)\cdot \tilde{X}_{\mu p, err, norm}(s)$ with $\tilde{X}_{\mu p, err, norm}(s) = \tilde{X}_{\mu p, err}(s)/\tilde{X}_{\mu p, max}$. Lastly we define a weight function to enforce the actuator limits of the micro manipulator of $\dot{\tilde{x}}_{\mu a,max} = \unitfrac[0.1]{m}{s}$ using a static gain of $K_{\dot{x}} = W_{\dot{x}} = 1/\dot{\tilde{x}}_{\mu a,max} = 10$. The last performance function is defined as $p_{\dot{x}}(s) = W_{\dot{x}}\cdot \dot{\tilde{X}}_{\mu a,des}(s)$.

\pagebreak
\section{Experimental Results}
\label{sec:experimentalResults}
To underline the increase in dynamics of our proposed control architecture, we compare it to the state-of-the-art leader-follower-based macro-micro manipulation approach (LF) from \cite{friedrichHighlyDynamicPhysical2025} and a traditional robot-based force-torque-control (RB) using only the macro manipulator and a force-torque sensor. For that, we conduct various experiments to validate the different properties. First, we conduct experiments to identify the control bandwidth (\ref{sec:expIdentificationBandwidth}) and metrics to get insights into the force behavior and contact establishment (\ref{sec:expCollision}) as well as the task execution time, compare Fig. \ref{fig:experiments}. Second, we validate our architecture in different real-world applications, like peg-in-hole, gear and circuit board assembly (\ref{sec:expAssembly}). The used state of the art LF-based architecture consists of two controllers, one damping controller for the force control and a PD-controller for the position. The force controller commands the micro manipulator, while the position controller commands the macro to ensure that the micro manipulator stays in its center position. For the LF architecture we were not able to use the same performance functions described in the previous section. This architecture is not able to use the macro manipulator for the compensation of the micros position if no force overshoot is allowed. To tune the parameters we proceeded as follows: In the first step the gains of the position controller are set to zero and the gain of the force controller is increased until the step response shows a force overshoot. In the second step we increased the P and D gain until a force overshoot of \unit[10]{\%} is reached. For the control parameters of the RB architecture, we only use the performance function $p_F$.

The control algorithms are implemented in MATLAB/Simulink on a Real-Time Target PC. For the communication with the robot (FANUC LR Mate 200iD/7L) we use a custom ROS FANUC driver. 
\begin{figure}[!h]
\centering
\def\svgwidth{\linewidth}
\hspace*{0.15cm}
\begingroup%
  \makeatletter%
  \providecommand\color[2][]{%
    \errmessage{(Inkscape) Color is used for the text in Inkscape, but the package 'color.sty' is not loaded}%
    \renewcommand\color[2][]{}%
  }%
  \providecommand\transparent[1]{%
    \errmessage{(Inkscape) Transparency is used (non-zero) for the text in Inkscape, but the package 'transparent.sty' is not loaded}%
    \renewcommand\transparent[1]{}%
  }%
  \providecommand\rotatebox[2]{#2}%
  \newcommand*\fsize{\dimexpr\f@size pt\relax}%
  \newcommand*\lineheight[1]{\fontsize{\fsize}{#1\fsize}\selectfont}%
  \ifx\svgwidth\undefined%
    \setlength{\unitlength}{324.95999908bp}%
    \ifx\svgscale\undefined%
      \relax%
    \else%
      \setlength{\unitlength}{\unitlength * \real{\svgscale}}%
    \fi%
  \else%
    \setlength{\unitlength}{\svgwidth}%
  \fi%
  \global\let\svgwidth\undefined%
  \global\let\svgscale\undefined%
  \makeatother%
  \begin{picture}(1,0.68906942)%
    \lineheight{1}%
    \setlength\tabcolsep{0pt}%
    \put(0,0){\includegraphics[width=\unitlength,page=1]{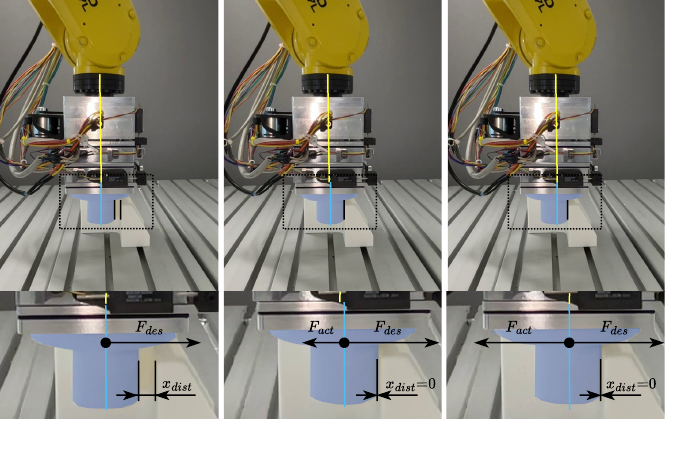}}%
    \put(0.14941582,0.02882854){\color[rgb]{0,0,0}\makebox(0,0)[lt]{\lineheight{1.25}\smash{\begin{tabular}[t]{l}(a)\end{tabular}}}}%
    \put(0.47891312,0.03104412){\color[rgb]{0,0,0}\makebox(0,0)[lt]{\lineheight{1.25}\smash{\begin{tabular}[t]{l}(b)\end{tabular}}}}%
    \put(0.80965668,0.03104412){\color[rgb]{0,0,0}\makebox(0,0)[lt]{\lineheight{1.25}\smash{\begin{tabular}[t]{l}(c)\end{tabular}}}}%
  \end{picture}%
\endgroup%

\caption{Procedure for the experiments. At the start of a experiment we have a distance $x_{dist}$ between the end-effector and a 3D-printed block. We command a desired force $F_{des}$ and measure the resulting force $F_{act}$ to the object in contact. The yellow line indicates the TCP and the blue line marks the micro manipulators center position. (a) Starting position. (b) Contact with the object and approaching $F_{des}$. (c) $F_{des} = F_{act}$ and micro in its center position.} 
\label{fig:experiments}
\end{figure}
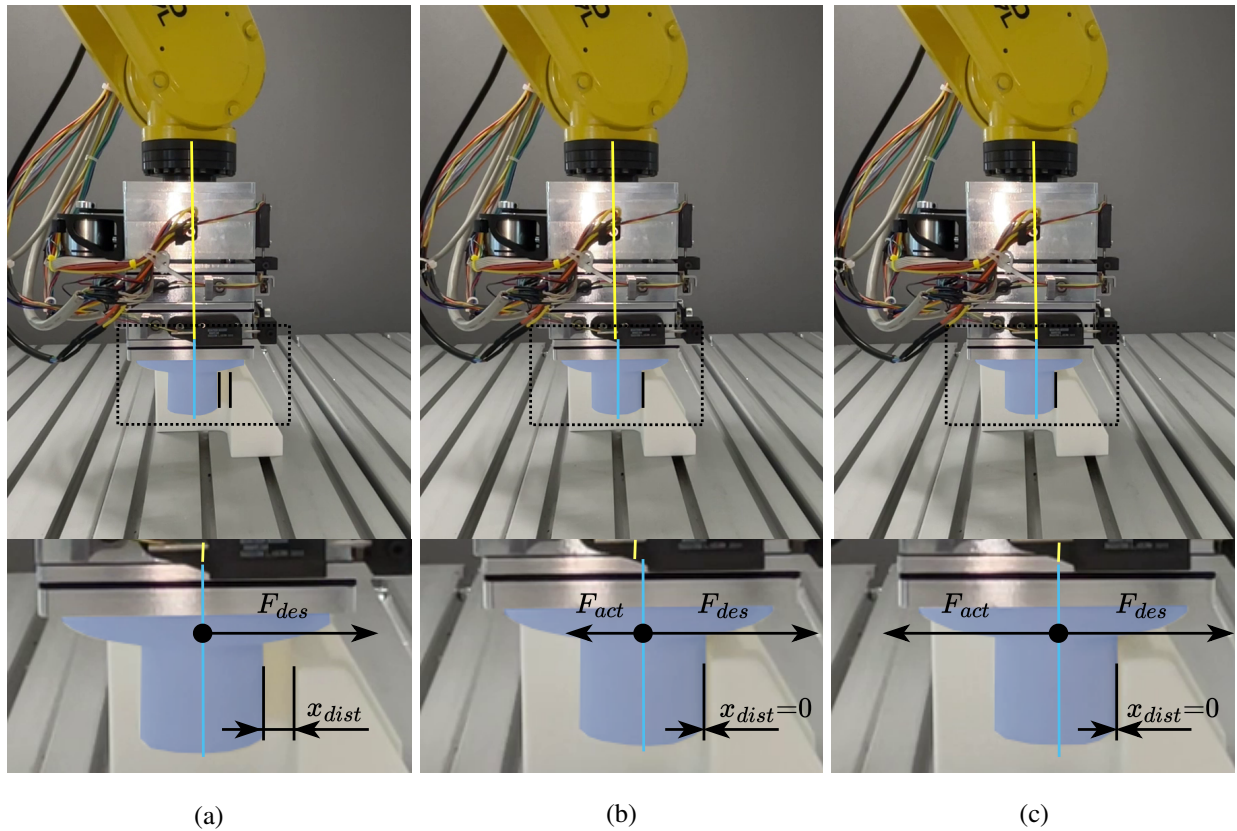\pagebreak
\subsection{Identification of the force control bandwidth}\label{sec:expIdentificationBandwidth}
\begin{figure}[!h]
\centering
\includegraphics[width=\linewidth]{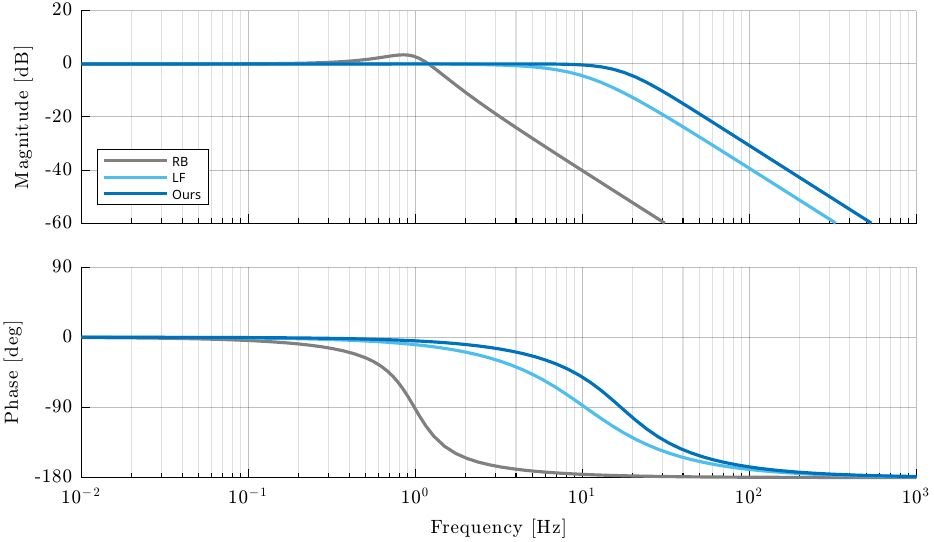}
\caption{Bode plot for the fitted transfer functions of the closed force control loop with input $F_{des}$ and output $F_{act}$.}
\label{fig:forceTrajectoryChirp}
\end{figure}

Similar to the identification of the surrogate models, we use a sine sweep with $f_{id,F} = [0.1\dots 50]\;\unit{Hz}$, a amplitude of $A_F = \unit[5]{N}$ and a force offset of $F_{offset} = \unit[15]{N}$ as the input signal during contact with an object. We measure the resulting force $F_{act}$ and fit transfer functions for a reduced order model to the data. This reduced order model is able to reflect the systems behavior. The resulting transfer functions for each approach are shown in Fig. \ref{fig:forceTrajectoryChirp}. The $\unit[-3]{dB}$ cutoff frequency is at $\unit[\expforceTrajectoryChirparccvIICutoffResVal]{Hz}$ (Ours), $\unit[\expforceTrajectoryChirparccvICutoffResVal]{Hz}$ (LF) and $\unit[\expforceTrajectoryChirprobotCutoffResVal]{Hz}$ (RB). Our approach outperforms the LF-based and RB-based approach in closed loop force control by a factor of $2.1$ and $12.5$, respectively.

\subsection{Collision with an object}\label{sec:expCollision}
For the second and third experiment we have a starting position of the macro-micro manipulator with a distance $x_{dist}$ to an object, see Fig. \ref{fig:experiments}. The object is a 3D-printed block mounted to the robot table, see Fig. \ref{fig:experiments}. The desired force is set to $F_{des} = \unit[20]{N}$ in the direction of the object. We carry out two different experiments. The first, where the collision is inside the ROM of the micro manipulator, and the second outside the ROM. The second one allows, above all, the evaluation of the shared control between the macro and micro manipulators.

\subsubsection{Collision inside the micro manipulators ROM}\label{sec:expCollisionInsideROM}
\begin{figure}[!h]
\centering
\includegraphics[width=\linewidth]{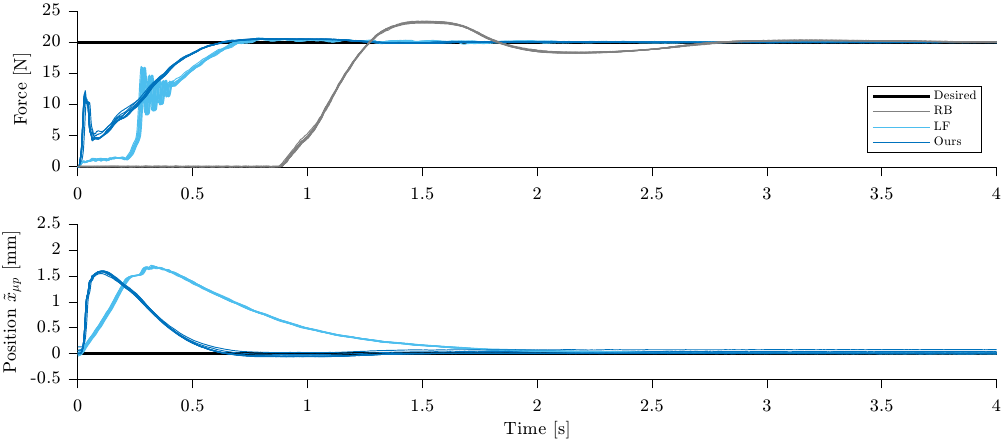}
\caption{Results from collision inside ROM experiment. The starting position is at a distance of $\unit[2]{mm}$ to the object. For each control approach the experiment is carried out ten times. The jittering is caused by inertia and the transition between the local models for the state in contact and not in contact.}
\label{fig:collisionXInsideROM}
\end{figure}

The starting position for the second experiment is at a distance of $x_{dist} = \unit[2]{mm}$, which is inside the micros ROM of $\pm\unit[2.5]{mm}$. Each experiment is carried out ten times. The results are shown in Fig. \ref{fig:collisionXInsideROM}. The time to reach the object is $\unit[\expCollisionInsideROMarccvIIContactReachedResVal]{s}$ (Ours), $\unit[\expCollisionInsideROMarccvIContactReachedResVal]{s}$ (LF) and $\unit[\expCollisionInsideROMrobotContactReachedResVal]{s}$ (RB) and it takes $\unit[\expCollisionInsideROMarccvIIForceReachedResVal]{s}$ (Ours), $\unit[\expCollisionInsideROMarccvIForceReachedResVal]{s}$ (LF) and $\unit[\expCollisionInsideROMrobotForceReachedResVal]{s}$ (RB) to reach the desired force. The force tracking RMSE is $\unit[\expCollisionInsideROMarccvIIRMSEForceResVal]{N}$ (Ours), $\unit[\expCollisionInsideROMarccvIRMSEForceResVal]{N}$ (LF) and $\unit[\expCollisionInsideROMrobotRMSEForceResVal]{N}$ (RB). The time to reach the micro manipulators desired position $\tilde{x}_{\mu p,des}$ after reaching the desired force is \unit[\expCollisionInsideROMarccvIIZeroPosResVal]{s} for our approach and \unit[\expCollisionInsideROMarccvIZeroPosResVal]{s} for the LF-based approach. This significant lower position time allows safety guarantees, especially when higher force need to be absorbed.

\subsubsection{Collision outside the micro manipulators ROM}\label{sec:expCollisionOutsideROM}
\begin{figure}[!h]
\centering
\includegraphics[width=\linewidth]{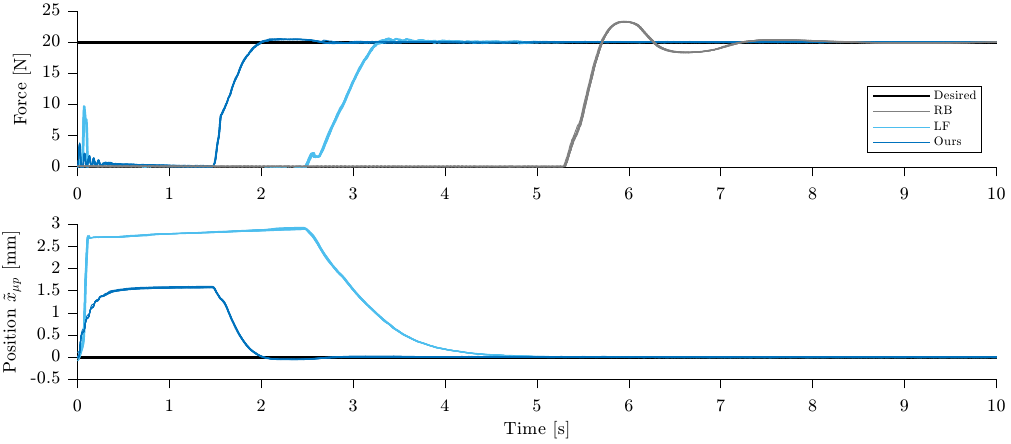}
\caption{Results from collision outside ROM experiment. The starting position is at a distance of $\unit[10]{mm}$ to the object. For each control approach the experiment is carried out ten times. The jittering is caused by inertia and the transition between the local models for the state in contact and not in contact.}
\label{fig:collisionXOutsideROM}
\end{figure}

For the third experiment the distance to the object is set at $\unit[10]{mm}$, which is well outside the ROM of the micro manipulator. Again, each experiment is carried out ten times. The results are shown in Fig. \ref{fig:collisionXOutsideROM}. The time to reach the object is $\unit[\expCollisionOutsideROMarccvIIContactReachedResVal]{s}$ (Ours), $\unit[\expCollisionOutsideROMarccvIContactReachedResVal]{s}$ (LF) and $\unit[\expCollisionOutsideROMrobotContactReachedResVal]{s}$ (RB) and it takes $\unit[\expCollisionOutsideROMarccvIIForceReachedResVal]{s} $ (Ours), $\unit[\expCollisionOutsideROMarccvIForceReachedResVal]{s}$ (LF) and $\unit[\expCollisionOutsideROMrobotForceReachedResVal]{s}$ (RB) to reach the desired force. The force tracking RMSE is $\unit[\expCollisionOutsideROMarccvIIRMSEForceResVal]{N}$ (Ours), $\unit[\expCollisionOutsideROMarccvIRMSEForceResVal]{N}$ (LF) and $\unit[\expCollisionOutsideROMrobotRMSEForceResVal]{N}$ (RB). The time to reach the micro manipulators desired position $\tilde{x}_{\mu p,des}$ after reaching the desired force is $\unit[\expCollisionOutsideROMarccvIIZeroPosResVal]{s}$ for our approach and $\unit[\expCollisionOutsideROMarccvIZeroPosResVal]{s}$ for the LF-based approach.

\subsection{Following of a force trajectory}\label{sec:expForceTrajectory}
\begin{figure}[!h]
\centering
\includegraphics[width=\linewidth]{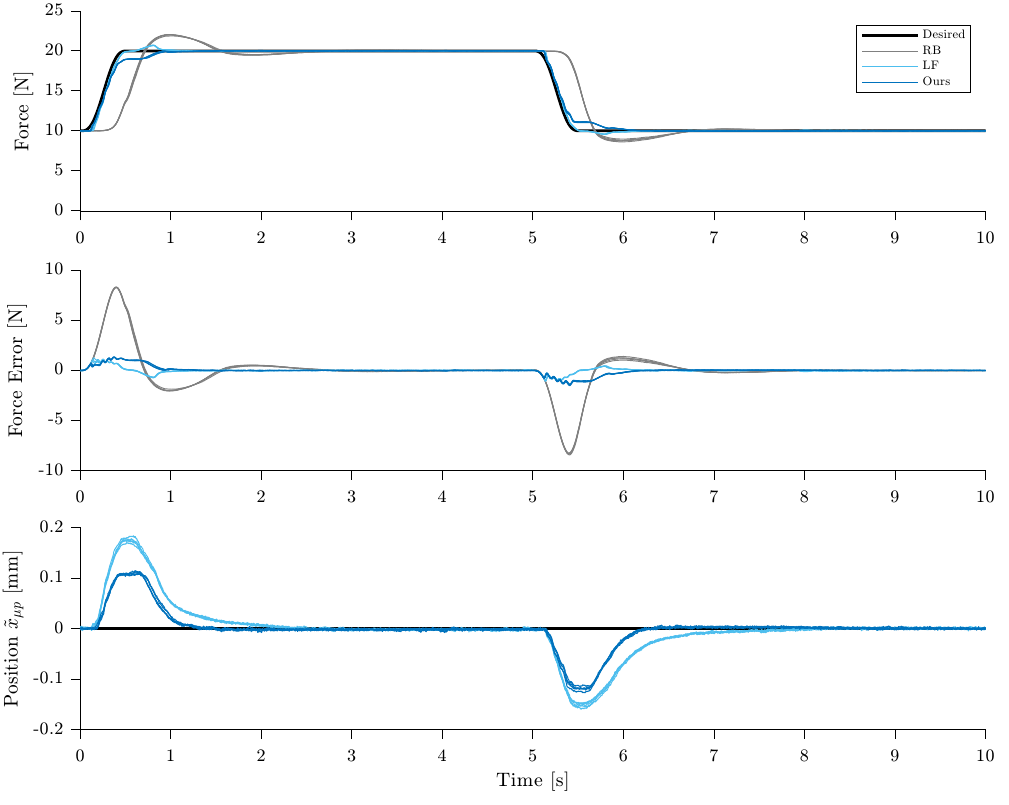}
\caption{Results from the force trajectory following experiment. The desired force is switched between \unit[10]{N} and \unit[20]{N} using a fifth order polynomial.}
\label{fig:forceTrajectoryRIR}
\end{figure}

For this experiment we follow a force trajectory switching between $F_{des} = \unit[10]{N}$ and $\unit[20]{N}$ using a fifth order polynomial to interpolate the switching trajectory following behavior. Each experiment is carried out five times using the experimental setup from Fig. \ref{fig:experiments}. The results are shown in Fig. \ref{fig:forceTrajectoryRIR}. The maximum force error is \unit[\expforceTrajectoryRIRarccvIIMaxErrorResVal]{N} (Ours), \unit[\expforceTrajectoryRIRarccvIMaxErrorResVal]{N} (LF) and \unit[\expforceTrajectoryRIRrobotMaxErrorResVal]{N} (RB). Our approach shows a slightly higher maximum force error and tracking error with a RMSE of \unit[\expforceTrajectoryRIRarccvIIRMSEForceResVal]{N} compared to \unit[\expforceTrajectoryRIRarccvIRMSEForceResVal]{N} for the LF-based approach. The cause for this is that in our proposed architecture the force controller and position controller work against each other while there are changes in the desired force. However, using our approach there is no overshoot of the desired force. The tracking RMSE for the RB-based approach is \unit[\expforceTrajectoryRIRrobotRMSEForceResVal]{N}.

\pagebreak
\subsection{Assembly tasks}\label{sec:expAssembly}
\begin{figure*}[hb!]
\centering
\includegraphics[width=\textwidth, keepaspectratio]{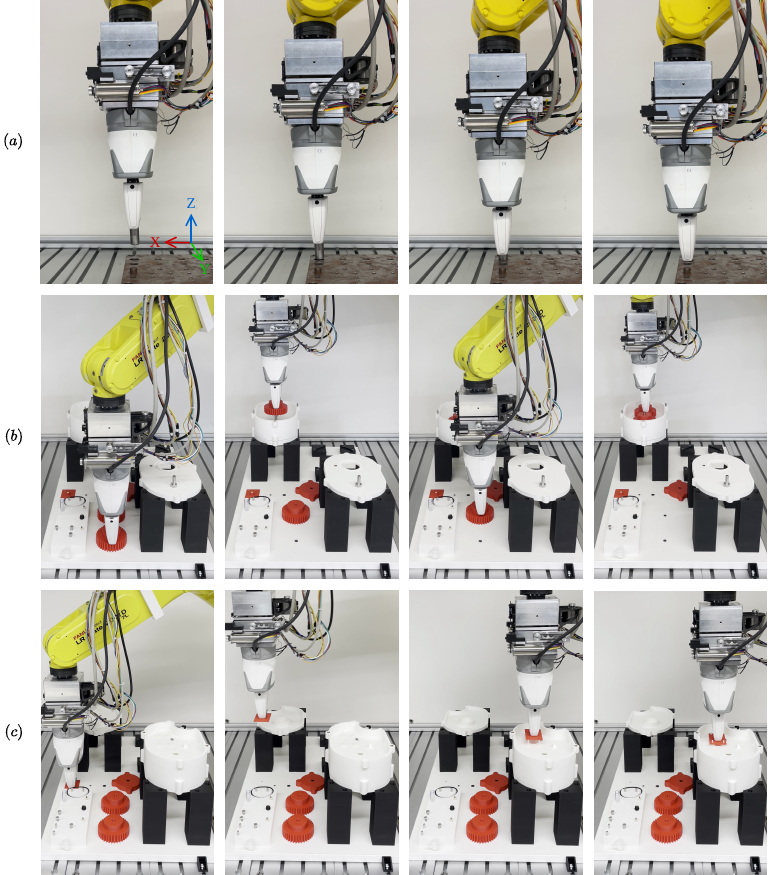}
\caption{Series of images of the performed assembly tasks. (a) Peg-in-hole experiment. Gear assembly task (b) and circuit board installation (c) from the assembly benchmark for industrial tasks \cite{schemppLABITLongHorizonRobotic2026}.}
\label{fig:resultsSequence}
\end{figure*}

To underline the effectiveness of our proposed control architecture in real-world applications, we conduct different assembly experiments. The setups for the peg-in-hole, gear assembly and circuit board assembly are shown in Fig. \ref{fig:resultsSequence} (a), \ref{fig:resultsSequence} (b) and \ref{fig:resultsSequence} (c), respectively.
\vspace*{-0.1cm}

\subsubsection{Peg-in-hole}\label{subsec:expPiH}
For the first assembly task we chose a traditional peg-in-hole application, see Fig. \ref{fig:resultsSequence} (a), since this is the most common type of task in assembly processes. The peg has a diameter of $\unit[16]{mm}$ for a hole of diameter $\unit[16.8]{mm}$, leaving a tolerance of $\pm\unit[0.4]{mm}$. The peg is manually inserted into the gripper, leading to a slight uncertainty of the pins pose. The desired force is set to $\unit[-20]{N}$ for the z-axis and $\unit[0]{N}$ for the x- and y-axis. Each experiment is performed ten times. The mean time for successful insertion is $\unit[\exppiharccvIItimeResVal]{s}$ (Ours), $\unit[\exppiharccvItimeResVal]{s}$ (LF) and $\unit[\exppihrobottimeResVal]{s}$ (RB).

\subsubsection{Gear assembly}\label{subsec:expGear}
The last two experiments are assembly tasks from the assembly benchmark for industrial tasks \cite{schemppLABITLongHorizonRobotic2026}. The first task involves installing two gears into a housing, compare Fig. \ref{fig:resultsSequence} (b). For this the robot approaches a initial position above the respective gear shaft and switches to the respective force controller to install the gear. The pickup and drop-off positions of the gears introduce uncertainty to the process that is compensated by the force controllers. Similar to the peg-in-hole application, the desired force is set to $\unit[-20]{N}$ for the z-axis and $\unit[0]{N}$ for the x- and y-axis. The desired position of the micro manipulator $\tilde{x}_{\mu p, des}$ follows a spiral search trajectory in order to find the gear shaft and to align the gear teeth. The assembly process is repeated three times for each control approach and the best result is used. The whole assembly process, i.e. gripping and moving objects under position control and installing the objects using force control, is completed in $\unit[\expgeararccvIItimeAssemblyResVal]{s}$ (Ours), $\unit[\expgeararccvItimeAssemblyResVal]{s}$ (LF) and $\unit[\expgearrobottimeAssemblyResVal]{s}$ (RB). The subtask of installing the first gear under force control takes $\unit[\expgeararccvIItimeGearIResVal]{s}$ (Ours), $\unit[\expgeararccvItimeGearIResVal]{s}$ (LF) and $\unit[\expgearrobottimeGearIResVal]{s}$ (RB) and for the second gear $\unit[\expgeararccvIItimeGearIIResVal]{s}$ (Ours), $\unit[\expgeararccvItimeGearIIResVal]{s}$ (LF) and $\unit[\expgearrobottimeGearIIResVal]{s}$ (RB).

\subsubsection{Circuit board assembly}\label{subsec:expCircuitBoard}
For the second task of the assembly benchmark, we install a circuit board into the housing, see Fig. \ref{fig:resultsSequence} (c). This circuit board is clipped in by four pins. For this task we set the desired force to $\unit[-10]{N}$ for the z-axis to avoid damaging the pins. Similar to the gear assembly, the micro manipulator follows a spiral search trajectory to align the pins to the board. The assembly process is repeated three times for each control approach and the best result is used. The whole assembly process, i.e. gripping and moving objects under position control and installing the objects using force control, is completed in $\unit[\expboardarccvIItimeAssemblyResVal]{s}$ (Ours), $\unit[\expboardarccvItimeAssemblyResVal]{s}$ (LF) and $\unit[\expboardrobottimeAssemblyResVal]{s}$ (RB). Excluding the position controlled times, the times are $\unit[\expboardarccvIItimeBoardResVal]{s}$ (Ours), $\unit[\expboardarccvItimeBoardResVal]{s}$ (LF) and $\unit[\expboardrobottimeBoardResVal]{s}$ (RB).

\subsection{Overall comparison}
To further highlight the advantage of our proposed approach, we list the results of the experiments together with the relative improvement to the other two approaches in Tab. \ref{tab:overallComparison}. Our approach of actively incorporating the macro manipulator in the overall force control significantly outperforms the other approaches. The only exception for this is experiment \ref{sec:expForceTrajectory} where our approach shows a slightly higher maximum force error and RMSE for the force compared to the LF-based approach, which is caused by the effects described in (\ref{sec:expForceTrajectory}).
\pagebreak
\begin{table*}[h]
\caption{Results for all experiments and the relative improvement by using our approach.}
\label{tab:overallComparison}
\centering
\begin{tabular}{llclllcll}
\toprule
\multirow{2}{*}{Exp.} & \multirow{2}{*}{Metric} && \multicolumn{3}{c}{Result} && \multicolumn{2}{c}{Improvement vs.}\\
\cmidrule{4-6} \cmidrule{8-9}
 &&& ours & LF-based & RB-based && LF-based & RB-based \\
\midrule
\ref{sec:expIdentificationBandwidth} & \unit[-3]{dB} Cutoff frequency && $\boldsymbol{\unit[\expforceTrajectoryChirparccvIICutoffResVal]{Hz}}$ & $\unit[\expforceTrajectoryChirparccvICutoffResVal]{Hz}$ & $\unit[\expforceTrajectoryChirprobotCutoffResVal]{Hz}$ && $\unit[\expforceTrajectoryChirparccvICutoffRelImprov]{\%}$ & $\unit[\expforceTrajectoryChirprobotCutoffRelImprov]{\%}$ \\
\ref{sec:expCollisionInsideROM} & Contact established && $\boldsymbol{\unit[\expCollisionInsideROMarccvIIContactReachedResVal]{s}}$ & $\unit[\expCollisionInsideROMarccvIContactReachedResVal]{s}$ & $\unit[\expCollisionInsideROMrobotContactReachedResVal]{s}$ && $\unit[\expCollisionInsideROMarccvIContactReachedRelImprov]{\%}$ & $\unit[\expCollisionInsideROMrobotContactReachedRelImprov]{\%}$ \\
& RMSE force && $\boldsymbol{\unit[\expCollisionInsideROMarccvIIRMSEForceResVal]{N}}$ & $\unit[\expCollisionInsideROMarccvIRMSEForceResVal]{N}$ & $\unit[\expCollisionInsideROMrobotRMSEForceResVal]{N}$ && $\unit[\expCollisionInsideROMarccvIRMSEForceRelImprov]{\%}$ & $\unit[\expCollisionInsideROMrobotRMSEForceRelImprov]{\%}$ \\
& Desired force reached && $\boldsymbol{\unit[\expCollisionInsideROMarccvIIForceReachedResVal]{s}}$ & $\unit[\expCollisionInsideROMarccvIForceReachedResVal]{s}$ & $\unit[\expCollisionInsideROMrobotForceReachedResVal]{s}$ && $\unit[\expCollisionInsideROMarccvIForceReachedRelImprov]{\%}$ & $\unit[\expCollisionInsideROMrobotForceReachedRelImprov]{\%}$ \\
& Des. position $\tilde{x}_{\mu p}$ reached && $\boldsymbol{\unit[\expCollisionInsideROMarccvIIZeroPosResVal]{s}}$ & $\unit[\expCollisionInsideROMarccvIZeroPosResVal]{s}$ & $-$ && $\unit[\expCollisionInsideROMarccvIZeroPosRelImprov]{\%}$ & $-$\\
\ref{sec:expCollisionOutsideROM} & Contact established && $\boldsymbol{\unit[\expCollisionOutsideROMarccvIIContactReachedResVal]{s}}$ & $\unit[\expCollisionOutsideROMarccvIContactReachedResVal]{s}$ & $\unit[\expCollisionOutsideROMrobotContactReachedResVal]{s}$ && $\unit[\expCollisionOutsideROMarccvIContactReachedRelImprov]{\%}$ & $\unit[\expCollisionOutsideROMrobotContactReachedRelImprov]{\%}$ \\
& RMSE force && $\boldsymbol{\unit[\expCollisionOutsideROMarccvIIRMSEForceResVal]{N}}$ & $\unit[\expCollisionOutsideROMarccvIRMSEForceResVal]{N}$ & $\unit[\expCollisionOutsideROMrobotRMSEForceResVal]{N}$ && $\unit[\expCollisionOutsideROMarccvIRMSEForceRelImprov]{\%}$ & $\unit[\expCollisionOutsideROMrobotRMSEForceRelImprov]{\%}$ \\
& Desired force reached && $\boldsymbol{\unit[\expCollisionOutsideROMarccvIIForceReachedResVal]{s}}$ & $\unit[\expCollisionOutsideROMarccvIForceReachedResVal]{s}$ & $\unit[\expCollisionOutsideROMrobotForceReachedResVal]{s}$ && $\unit[\expCollisionOutsideROMarccvIForceReachedRelImprov]{\%}$ & $\unit[\expCollisionOutsideROMrobotForceReachedRelImprov]{\%}$ \\
& Des. position $\tilde{x}_{\mu p}$ reached && $\boldsymbol{\unit[\expCollisionOutsideROMarccvIIZeroPosResVal]{s}}$ & $\unit[\expCollisionOutsideROMarccvIZeroPosResVal]{s}$ & $-$ && $\unit[\expCollisionOutsideROMarccvIZeroPosRelImprov]{\%}$ & $-$\\
\ref{sec:expForceTrajectory} & Maximum force error && $\unit[\expforceTrajectoryRIRarccvIIMaxErrorResVal]{N}$ & $\boldsymbol{\unit[\expforceTrajectoryRIRarccvIMaxErrorResVal]{N}}$ & $\unit[\expforceTrajectoryRIRrobotMaxErrorResVal]{N}$ && $\unit[\expforceTrajectoryRIRarccvIMaxErrorRelImprov]{\%}$ & $\unit[\expforceTrajectoryRIRrobotMaxErrorRelImprov]{\%}$ \\
& RMSE force && $\unit[\expforceTrajectoryRIRarccvIIRMSEForceResVal]{N}$ & $\boldsymbol{\unit[\expforceTrajectoryRIRarccvIRMSEForceResVal]{N}}$ & $\unit[\expforceTrajectoryRIRrobotRMSEForceResVal]{N}$ && $\unit[\expforceTrajectoryRIRarccvIRMSEForceRelImprov]{\%}$ & $\unit[\expforceTrajectoryRIRrobotRMSEForceRelImprov]{\%}$ \\
& RMSE position $\tilde{x}_{\mu p}$ && $\boldsymbol{\unit[\expforceTrajectoryRIRarccvIIRMSEPosResVal{\cdot} 10^{\expforceTrajectoryRIRarccvIIRMSEPosResExp}]{m}}$ &  $\unit[\expforceTrajectoryRIRarccvIRMSEPosResVal{\cdot} 10^{\expforceTrajectoryRIRarccvIRMSEPosResExp}]{m}$ & $-$ && $\unit[\expforceTrajectoryRIRarccvIRMSEPosRelImprov]{\%}$ & $-$\\
\ref{subsec:expPiH} & Task time && $\boldsymbol{\unit[\exppiharccvIItimeResVal]{s}}$ & $\unit[\exppiharccvItimeResVal]{s}$ & $\unit[\exppihrobottimeResVal]{s}$ && $\unit[\exppiharccvItimeRelImprov]{\%}$ & $\unit[\exppihrobottimeRelImprov]{\%}$ \\
\ref{subsec:expGear} & Total assembly time && $\boldsymbol{\unit[\expgeararccvIItimeAssemblyResVal]{s}}$ & $\unit[\expgeararccvItimeAssemblyResVal]{s}$ & $\unit[\expgearrobottimeAssemblyResVal]{s}$ && $\unit[\expgeararccvItimeAssemblyRelImprov]{\%}$ & $\unit[\expgearrobottimeAssemblyRelImprov]{\%}$ \\
& Installation time gear 1 && $\boldsymbol{\unit[\expgeararccvIItimeGearIResVal]{s}}$ & $\unit[\expgeararccvItimeGearIResVal]{s}$ & $\unit[\expgearrobottimeGearIResVal]{s}$ && $\unit[\expgeararccvItimeGearIRelImprov]{\%}$ & $\unit[\expgearrobottimeGearIRelImprov]{\%}$ \\
& Installation time gear 2 && $\boldsymbol{\unit[\expgeararccvIItimeGearIIResVal]{s}}$ & $\unit[\expgeararccvItimeGearIIResVal]{s}$ & $\unit[\expgearrobottimeGearIIResVal]{s}$ && $\unit[\expgeararccvItimeGearIIRelImprov]{\%}$ & $\unit[\expgearrobottimeGearIIRelImprov]{\%}$ \\
\ref{subsec:expCircuitBoard} & Total assembly time && $\boldsymbol{\unit[\expboardarccvIItimeAssemblyResVal]{s}}$ & $\unit[\expboardarccvItimeAssemblyResVal]{s}$ & $\unit[\expboardrobottimeAssemblyResVal]{s}$ && $\unit[\expboardarccvItimeAssemblyRelImprov]{\%}$ & $\unit[\expboardrobottimeAssemblyRelImprov]{\%}$ \\
& Installation time circuit board && $\boldsymbol{\unit[\expboardarccvIItimeBoardResVal]{s}}$ & $\unit[\expboardarccvItimeBoardResVal]{s}$ & $\unit[\expboardrobottimeBoardResVal]{s}$ && $\unit[\expboardarccvItimeBoardRelImprov]{\%}$ & $\unit[\expboardrobottimeBoardRelImprov]{\%}$ \\
\bottomrule
\end{tabular}
\end{table*}

\section{Discussion and Outlook}
\label{sec:discussionAndOutlook}
We propose a novel control architecture for macro-micro manipulation by actively incorporating the macro manipulator in the interaction control. This leads to a significantly higher force control bandwidth compared to the commonly used leader-follower approach for macro-micro manipulation. The proposed surrogate models allow for efficient controller design and adaption to hardware changes. The performance improvement of the proposed approach is shown in the different experiments, especially the high bandwidth, which leads to drastically lower cycle times and allows more sensitive robot applications. For the future we plan to learn the environment behavior by visual observation to automatically adapt to different contact situations. In addition, we see a great benefit in incorporating visual data to estimate the time of contact, which can yield to further improvement in dynamics of the overall system.

\section*{Acknowledgments}
We thank Matthias Haag and Marco Santin for developing and providing the hardware for the experiments.
\section*{Funding sources}
This work was financed by the Baden-Württemberg Stiftung in the scope of the AUTONOMOUS ROBOTICS project ISASDeMoRo.
\section*{Author contributions: CRediT}
\textbf{Patrick Frank:} Conceptualization, Methodology, Software, Validation, Investigation, Data Curation, Writing - Original Draft, Visualization. 
\textbf{Christian Friedrich:} Conceptualization, Writing - Review \& Editing, Supervision, Project administration, Funding acquisition
\section*{Declaration of competing interests}
The authors declare that they have no known competing financial interests or personal relationships that could have appeared to influence the work reported in this paper.
\section*{Data availability}
Data will be made available on request.

\bibliographystyle{elsarticle-num}
\bibliography{ref}







\end{document}